\documentclass{article}
\usepackage[utf8]{inputenc}
\usepackage{apacite}
\usepackage{hyperref}
\usepackage{natbib}
\usepackage{graphicx}

\title{Paradigms of Computational Agency}
\author{Srinath Srinivasa \\ IIIT-Bangalore \\ \href{mailto:sri@iiitb.ac.in}{sri@iiitb.ac.in} \and Jayati Deshmukh \\ IIIT-Bangalore \\ \href{mailto:jayati.deshmukh@iiitb.org}{jayati.deshmukh@iiitb.org}}
\date{}

\begin{document}

\maketitle

\begin{abstract}
Agent-based models have emerged as a promising paradigm for addressing ever increasing complexity of information systems. In its initial days in the 1990s when object-oriented modeling was at its peak, an \textit{agent} was treated as a special kind of ``object'' that had a persistent state and its own independent thread of execution. Since then, agent-based models have diversified enormously to even open new conceptual insights about the nature of systems in general. This paper presents a perspective on the disparate ways in which our understanding of agency, as well as computational models of agency have evolved. Advances in hardware like GPUs, that brought neural networks back to life, may also similarly infuse new life into agent-based models, as well as pave the way for advancements in research on Artificial General Intelligence (AGI). \\

\end{abstract}

\section{Introduction}\label{sec:intro}

Today's information systems are complex, distributed, and need to scale over millions of users and a variety of devices, with guaranteed uptimes. As a result, top-down approaches for systems design and engineering are becoming increasingly infeasible. 

Starting sometime in the 1990s, a branch of systems engineering, has approached the problem of systemic complexity in a bottom-up fashion, by designing ``autonomous'' or ``intelligent'' agents that can proactively and autonomously act and decide on their own-- to address specific, local issues pertaining to their immediate requirements. They also can communicate and coordinate with one another to jointly solve larger problems. The autonomous nature of agents require some form of a rationale that justifies their actions. Given that, object-oriented modeling had attracted mainstream attention at that time, the distinction between mechanistic ``objects'' and autonomous ``agents'' were often summarized with this slogan~\citep{jennings1998roadmap}: \textit{Objects do it for free, agents do it for money}. 

Early research in agent-based systems focused on designing architectures, communication primitives, and knowledge structures for agents' reasoning. Several such independent research pursuits, also resulted in the emergence of standards organizations like FIPA\footnote{http://www.fipa.org/}, which is now an IEEE standards organization for promoting agent-based modeling and interoperability of its standards with other technologies~\citep{Poslad:2007:SPM:1293731.1293735}. 

But research interest soon moved from communication and coordination, to address the concept of agency itself. Agents are meant to take decisions ``autonomously''-- and the term ``autonomy'' needed sound conceptual and computational foundations. An autonomous agent needs to operate ``on its own'' and definitions for what this entails, distinguished different models of autonomy. Broadly, approaches to computational modeling of autonomy can be divided into the following research areas: \textit{normative}, \textit{adaptive}, \textit{quantitative}, and \textit{autonomic} models of agency. 

Normative models of agency, interpret agency as a combination of imperatives and discretionary entitlements. They also implement logical frameworks that encode different forms of individual and collective goals~\citep{castelfranchi1999deliberative,van2003towards,y2006normative}. Some normative elements for agents include: encoding of their goals, that in turn lead to encoding of their intentions or deliberative plans to achieve their goals, their belief about their environment, their obligations, their prohibitions, and so on. Interacting pairs of normative agents create \textit{contracts} that regulates their independent actions with respect to the others' actions. Systems of multiple, normative agents adopt collective \textit{deontics} or constitutions, that regulate overall behaviour~\citep{andrighetto2013normative}.

Adaptive frameworks for modeling agency, have emerged from problems where agents have to interact with complex and dynamic environments, like in autonomous driving and robotic navigation. These frameworks can either be model-driven where an underlying model of the environment is learned through interactions; or model-agnostic, where adaptations happen purely from positive or negative reinforcement signals from the environment~\citep{macal2005tutorial,shoham2003multi}.

The third paradigm of agency is based on quantitative methods based on decision theory and rational choice theory~\citep{ferber1999multi,parsons2002game,semsar2009multi}. These represent agents with a self-interest function, which then interact with their environment to obtain different kinds of payoffs resulting in a corresponding \textit{utility}. Rational agents then strive to make decisions in a way that results in \textit{utility maximization}. Rational choice is represented as pair-wise preference functions between choices, or as numerical payoffs. Interactions between agents are modeled as games representing \textit{confounded rationality}-- where rational choices of one agent may (positively or adversely) affect the prospects for others. 

A related stream of development, which we treat as the fourth paradigm of agency-- started somewhat independently from agent-based modeling, approaches agency by building a model of ``self.'' The field of autonomic computing~\citep{ganek2003dawning,kephart2003vision} first introduced by IBM, aimed to provide computational entities with self-management properties (also called ``self-*'' properties) like self-healing, self-tuning, self-recovery, etc. The field of Autopoiesis started by Maturana and Varela~\citep{maturana1991autopoiesis}, developed computational models of self-referential entities based on biological models of cognition. Yet another related field are those of Cybernetics and Artificial Life~\citep{johnston2008allure,komosinski2009artificial}, that addressed self-regulatory mechanisms that characterize natural life, into computational elements. Models developed here, were also used in the study of natural systems in evolutionary biology. 

\begin{figure}
    \centering
    \includegraphics[scale=0.4]{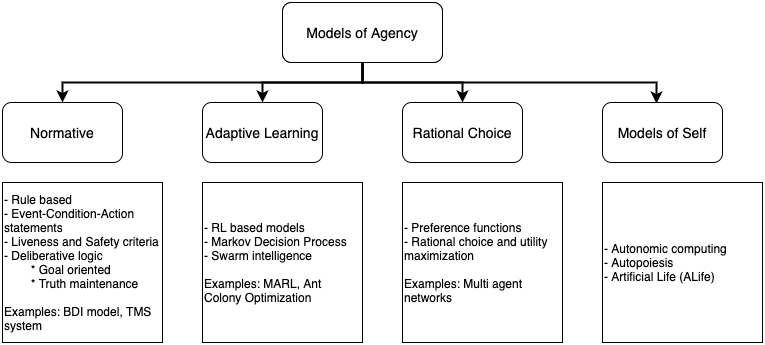}
    \caption{Paradigms of Computational Agency}
    \label{fig:CA_blockDiagram}
\end{figure}

In this chapter, we will organize our study of computational modeling of agency, along the four paradigms as shown in Figure \ref{fig:CA_blockDiagram}. We will look at the disparate viewpoints towards agency and the primary challenges addressed by them. 

\section{Normative Models of Agency}\label{sec:normative} 

In this approach, autonomy is defined in terms of rule-based specifications of discretionary and deliberative elements. 

Rule-based systems for autonomous decision-making may seem like a contradiction. That is, if an agent is dictated by rules, can it still be called autonomous? 

In the early days of agents research, rule-based approaches were adopted for agents to respond to various kinds of stimuli. These were called reflex agents~\citep{franklin1996agent}, whose rules were in the form of ECA (Event-Condition-Action) statements. One or more ECA rules would trigger in response to an external stimuli (Event), and based on the conditions that hold, the appropriate action would be performed. While these agents could display rich, adaptive behaviour, the rules and actions themselves had to be specified a priori. 

Subsequent research in normative models addressed a broader problem, to use rules to specify the boundaries within which autonomous decision-making takes place. Any system architecture, would need to have two forms of imperatives, which are mandatory. These are its \textit{liveness} and \textit{safety} criteria. These mandatory elements are specified by way of rules.

Liveness criteria, also called the set of ``obligations,'' represent properties that need to hold for the system to be considered functional. A property $p$ is said to be \textit{obligated}, represented as the modality $Op$ in deontic logic, if the system becomes inconsistent whenever $p$ does not hold. The term $Op$ is read as ``$p$ ought to be true,'' rather than ``$p$ is asserted to be true'' as with predicate logic statements. Similarly, safety criteria called ``prohibitions'' or ``forbidden'' properties are of the form $Fp$, which make the system inconsistent whenever they hold. 

It is important to note that, liveness and safety are not negations of one another. A property that is not obligated to hold, is not forbidden to hold; similarly, a property that is not forbidden to hold, is not obligated to hold. An agent that is not obligated to make a particular choice, is not forbidden from choosing it. Similarly, an agent that is not forbidden from choosing something, is not obligated to choose it.

Hence, the logic of norms has at least three modalities of truth. In between the obligated and forbidden regions, is the ``permitted'' or ``viable'' region~\citep{egbert2011quantifying,dignum2000towards,mukherjee2008validating} in which the agent can operate at its own discretion. 
Sometimes, the liveness criteria are also considered as part of the viable region. But here, we distinguish between the two because, upholding liveness properties are not subject to the agent's discretion. They are mandatory conditions, which the agents should necessarily uphold. 

The viable region is characterized by ``deliberative'' logic that guides autonomous decision-making by agents. Deliberative logic can in turn be broadly classified into two kinds: \textit{goal oriented} logic, and \textit{truth maintenance} logic. 

Goal oriented deliberative logic, represents autonomous decision-making in pursuit of a goal from a set of possible goals. One of the widely popular models for goal-oriented deliberative agents is the Belifs-Desires-Intention (BDI) model and its several variants~\citep{dignum2000towards,kinny1996methodology,meneguzzi2009norm,rao1991modeling,rao1995bdi}. 

This model comprises of three elements: 
\begin{description}
\item[Beliefs:] These represent the informational state of the agent, encoding its supposed knowledge about the environment and of other agents. Elements of an agent's belief may or may not be true, and can be revised on interaction with the environment.
\item[Desires:] These represent a set of goals that the agent wishes to perform. 
\item[Intentions:] These represent a (or a set of) goal(s) that the agent has committed to. An intention involves committing a goal to a set of actions, by choosing a plan from a set of available plans.
\end{description}

The creation of plans itself is not part of the BDI model, and is relegated to either human planners or a planning application. The choice of intention may be driven by several factors which involve one-shot, rational decision-making, learning, etc. Some BDI models also incorporate events that trigger activity in the agents-- like pursuing a goal or updating its beliefs. 

In contrast to goal oriented deliberative logic, truth maintenance systems (TMS)~\citep{doyle1977truth,huhns1991multiagent,mcallester1990truth} require agents to act autonomously to maintain one or more properties in the viable region, while interacting with external, uncertain environments. The generic model of a TMS comprises of the following: a property (or a set of properties) $\Phi$ which need to be maintained, a set of \textit{constraints}-- typically the liveness and safety constraints, and a set of \textit{premises} $\Sigma$, which represent a set of knowledge or belief elements about the state of the world, based on external interactions. The objective of the TMS is to compute entailments to establish whether $\Sigma \rightarrow \Phi$ holds.  

If the assertion $\Sigma \rightarrow \Phi$ can be proven to hold, then the TMS can maintain the required properties $\Phi$ as well as provide a justification for its maintenance-related actions. If $\Sigma \rightarrow \Phi$ can be proven to be false, then the TMS would need to perform corresponding actions such that $\Sigma$ no longer holds, and is replaced by another set of premises $\Sigma'$, which can entail $\Phi$. 

For example, consider an aircraft where $\Phi$ represents the altitude that needs to be maintained. $\Sigma$ represents the premise derived from the set of all input data like air speed, attitude, bank angle, drag, etc. If the premises can entail $\Phi$ it means that the current state of the aircraft can support maintenance of the altitude. If on the other hand, $\Sigma$ can be shown to not entail $\Phi$ it means that corrective action needs to be taken to adjust the aircraft state itself, so that the altitude can be maintained. 

There is a third case where the premises can neither entail the property that needs to be maintained, nor entail the negation of the property. In such cases, it is unknown to the agent whether $\Phi$ can be maintained. Such cases require TMS to employ non-monotonic and/or auto-epistemic elements to update its system of entailment rules.

Unlike goal-oriented logics, truth maintenance logics need to run continuously, to check current premises, entail required properties, and/or perform belief revision or non-monotonic updates to deal with uncertainty. Truth maintenance does not end with a single entailment computation.

\section{Adaptive Learning Based Models}\label{sec:adaptive} 

Adaptive learning based models are used in applications where agents have to interact with complex, dynamic environments; and need to continuously respond to changes. Examples include autonomous driving, robotic navigation, stochastic scheduling, swarm robotics, ant colony optimization, etc. 


Adaptive learning agents are modeled using reinforcement learning~\citep{sutton1998introduction}, which are in turn typically modeled as a Markov Decision Process (MDP). An MDP is characterized by a set of states $S$, and a set of actions $A$, and associated probability of state transitions for any given action. A term of the form $P_a(s,s')$ denotes the probability of the MDP state to transition from $s$ to $s'$ on performing action $a$. Any action by the agent may change the state of the interaction, and may also provide a feedback (sometimes called the ``reward'') from the environment, that could be either positive or negative. This is denoted as $R_a(s,s')$ indicating the reward on reaching $s'$ from $s$, by performing action $a$. 

Reinforcement learning addresses two forms of underlying challenges: the ``exploration vs exploitation'' dilemma, and the ``lookahead'' dilemma.  


The \textit{exploration vs exploitation} dilemma involves deciding between choosing the action with the best expected payoff at any given interaction state; or choosing a new action to explore more of the interaction state space. The \textit{look ahead} dilemma involves deciding whether to choose an action based on its immediate expected reward, or consider longer term prospects of having chosen such an action. Different reinforcement learning heuristics exist for addressing both the dilemmas. 

For finite MDPs, a well known algorithm called Q-learning~\cite{watkins1992q}, based on power iterations, is widely used to compute strategic payoffs based on unbounded look ahead. 

Reinforcement learning is conventionally used for single agent interactions with its environment. Other agents are considered to be part of the complex, dynamic environment that the given agent interacts with. 

However, the extension from a single-agent RL to a multi-agent RL problem is not straightforward. A shared space with several agents can thus be thought of as independent RL runs by each agent separately. This however, is known to be ineffective due to agents overfitting their best responses to each other's behaviours~\citep{lanctot2017unified}. Multi-Agent Reinforcement Learning (MARL) models were hence developed based on concepts of joint policy correlation between agents, where policies generated using deep reinforcement learning, are evaluated using game theoretic principle~\citep{bucsoniu2010multi,shoham2003multi}. With finite state spaces, Q-learning approaches were also extended to multi-agent systems~\citep{claus1998dynamics}. It is also seen that it is harder to design systems of joint learners as compared to independent learners, and while independent learners overfit to each other's behaviours, joint learning often don't perform significantly better as they become entrenched in local minima.

A related area of research is adaptive social learning agents. These are adaptive agents operating in a shared state space, that not only respond to feedback from the environment, but also interact with other agents either competitively or cooperatively, and may also share instantaneous information, and episodic and general knowledge~\citep{littman1994markov,tan1993multi}. 

Swarm intelligence~\citep{kennedy2006swarm,bonabeau1999swarm,beni2004swarm} is another direction of adaptive learning based models which is motivated by nature. It models a system of agents which act as a group without any centralized control. A variant of Swarm intelligence, called Ant Colony Optimization~\citep{dorigo1999ant} has been useful in context of multi-agent systems. It has been used in a variety of use-cases like resource-constrained project scheduling~\citep{merkle2002ant} and optimization in continuous domains~\citep{socha2008ant}.

\section{Rational Choice Based Models}\label{sec:rationalchoice} 

In this model of agency, autonomous agents are modeled as rational maximizers, driven by a self-interest function, and operating towards utility maximization. Mathematical underpinnings of such models derive from rational choice theory, decision theory, and game theory. 

Given a set of elements or actions $A$, classical rational choice theory going back to the works of von Neumann and Morgenstern~\citep{von2007theory}, defines the following preference functions between pairs of elements of $A$: $\prec$ (strong preference), $\preceq$ (weak preference), and $\parallel$ (indifference). 

Mechanisms for converting pairwise preference functions into quantitative payoffs are also provided, which are based on the concept of expected utility. Any given choice $L$ is represented as a set of pairs of the form $(a_i, p_i)$, where $a_i \in A$ represents one of the elements, and $p_i$ is the probability of receiving $a_i$. For any pair of choices $L$ and $M$, a quantitative payoff function $u()$ can be formulated such that $L \prec M$ iff $E(u(L)) < E(u(M))$, where $E()$ is the expected value of the payoff function based on the elements of the corresponding choices. 

Rational choice and game theoretic formalisms have been widely employed in designing agent-based systems~\citep{boella2001game,hogg1997socially,kraus1997negotiation,panait2005cooperative,parsons2002game}. This paradigm of agent-based modeling has been found to be particularly attractive for applications involving simulation and gamification for policy design, where the agents represent human stakeholders~\citep{parker2003multi,pan2007multi,schreinemachers2006land}. Rational choice theory and game theory have a long history of being used as mathematical underpinnings for human decision-making and behavioural economics-- and agent-based simulation models offer an attractive opportunity to model and simulate the repercussions of policy changes. 

Human rationality is known to deviate considerably from the classical model of rational choice. In addition to rational maximization, human rationality is characterized by factors like consideration for empathy and fairness, risk aversion, bounded rationality, and a variety of cognitive biases. Agent-based modeling have addressed these in various ways in order to simulate human behaviour and its emergent consequences more accurately~\citep{deshmukh2015evolution,kant2006modeling,manson2006bounded,santos2016dynamics,vidal1995recursive}. These extensions to classical rational choice models are important in simulating probable outcomes in emergency situations (like fire evacuations, for example), involving humans~\cite{pan2005multi,tang2008agent}. 

While rational choice theory is used to direct the behaviour of individual agents, this is insufficient when multiple agents have to operate in a shared state space. Interactions between disparate agents in a shared state space, can be broadly seen as either non-cooperative or cooperative, in nature. Correspondingly, these kinds of interactions derive theories from non-cooperative game theory and negotiation theory for the former~\citep{chakraborti2015statistical,gotts2003agent,tennenholtz1999electronic}, and from cooperative game theory and allocation theory for the latter forms of interaction~\citep{adler2002cooperative,albiero2007cooperative}. 

A related application area that have used the rational choice paradigm for modeling agents, is multi-agent networks. These applications study complex networks, by combining network science, rational choice theory and other related areas like evolutionary algorithms, to study different kinds of emergent properties arising from agents acting rationally in a networked environment~\cite{mei2015complex,patil2009breeding,patil2009classes,villez2011resilient}. Some indicative problems addressed by multi-agent networks include: constrained negotiation and agreement protocols~\citep{nedic2010constrained,meng2013event}, modeling diffusion and synchrony~\cite{faber2010exploring,kiesling2012agent,kim2011agent}, etc.

\section{Models of Self and Agency}\label{sec:autonomic} 

Lastly, we review literature from related fields that developed independently of agent-based modeling. All of these fields have tried to model the concept of ``self'' in computational entities, which is becoming increasingly relevant in agent-based systems as well. 

The field of \textit{autonomic computing} became a major area of research, after IBM coined this term in 2001, to denote self managing systems~\cite{computing2006architectural}. These include database backed information systems that could configure, protect, tune, heal and recover from failures on their own. Subsequently, autonomic computing has been pursued in various forms~\cite{kephart2003vision,kephart2005research,huebscher2008survey}. The main motivation (from natural self-governing systems) of building autonomic systems was to have systems which can manage themselves instead of having a team of skilled workforce to manage the system. The four main principles of self-management~\citep{kephart2003vision} are \textit{self-configuration}: systems which can configure its components, \textit{self-optimization}: systems which keep improving over time, \textit{self-healing}: systems which can diagnose and rectify its problems and \textit{self-protection}: systems which can defend itself from attacks. 

Autonomic architecture is used to design autonomic computing systems. It creates a network of autonomous elements, each managing its own internal state and interacting with other elements as well as the external environment. 

On similar lines, taking inspiration from cell biology, Maturana and Varela~\citep{varela1974autopoiesis,maturana1991autopoiesis} coined the term \textit{autopoiesis}, representing systems which can sustain itself without any external interaction. These systems determine its own structure in order to sustain in an environment. Autopoeisis has since then been extended by designing computational models~\citep{mcmullin2004thirty,di2005autopoiesis} in context of social autopoietic systems~\citep{seidl2004luhmann}.  

Advancements in biology and computational science has lead to the development of a new area at its intersection called \textit{Artificial Life} or \textit{ALife}~\citep{langton1997artificial,bedau2003artificial,aguilar2014past,bedau2000open}. It is used to model natural life and its associated processes using computational models. It can be used to analyze things like evolution and dynamics in natural systems. ALife models are classified as \textit{soft}: involving simulations of systems, \textit{hard}: involving hardware implementations (using robots) or \textit{wet}: involving biochemical synthesis of elements. 

All these autonomic models like autopoesis, artificial life, autonomic computing etc can be considered as the foundational models for designing autonomous systems. It can provide an initial framework to build systems of agents having agency, autonomy as well as self-interest. It is going to be even more relevant with the recent advancements in areas like self-driving cars or autonomous drones.

\section{Agents and AGI}\label{sec:agi} 

The ultimate dream of artificial intelligence (AI) research is to create computational models of ``general'' intelligence, or AGI. AGI, also called ``strong'' or ``full'' AI, is contrasted with ``weak'' or ``narrow'' AI that is built for specific applications. A basic ingredient of general AI is the need for ``common sense'' form of intelligence, that is adaptive and applicable in different contexts, and each contextual experience enhances its overall intelligence. 

Architectures for AGI have explored different pardigms. The Novamente AGI engine~\citep{goertzel2004novamente,goertzel2007novamente}, incorporates several paradigms of narrow AI including reinforcement learning, evolutionary algorithms, neural networks and symbolic computing into an underlying model of mental processes based on complex systems theory. Approaches like Universal AI~\citep{hutter2001towards} and G\"odel Machines~\citep{schmidhuber2007godel} develop self-rewriting systems that can completely reprogram themselves subject only to computability constraints. Architectures like SOAR~\citep{laird2008extending,laird2012soar,young1999soar} and ACT-R~\citep{anderson1996act} incorporates several elements of human cognition like semantic and episodic memory, working memory, emotion, etc. from elmentary building blocks, to create an architecture for generic problem-solving. The ARS architecture~\cite{Schaat2014ARSAA} aims to develop an agent-based model of the human mind simulated as an artificial life engine, to explore general intelligence. 

The recent resurgence of interest in AI has been brought about by advances in parallel computing architectures like Graphics Programming Units (GPUs) that enabled implementation of large artificial neural networks (ANNs). This led to the field of deep learning where ANNs could even detect features automatically, and perform several forms of perception, recognition and linguistic functions. 

However, it is widely recognized that an artificial neuron does not represent how a natural neuron works. An artificial neuron is modeled as a gate, where an activation function is triggered by values on its several input lines. The gate metaphor of the foundations of computation has its roots in electrical engineering. However, natural neurons and other building blocks of life (muscles, cells, tissues, etc.) are known to be autonomous decision-makers~\cite{moreno2005agency} rather than passive gates. Agency in nature seems to be a balancing act between autonomous entities striving to sustain themselves, and explore or interact with their environment. 

General intelligence hence needs to be more of a truth maintenance system, rather than as a goal-oriented system. Preferences defining self interest, as well as declarative elements of one's knowledge are in turn rooted in considerations of sustainability of one's self and interaction with the environment. 

\section{Conclusions}\label{sec:conclusion}

In this chapter, we looked at how models of computational agency have evolved over time. Initially, agents were designed as normative elements comprising of discretionary and imperative regions. Although agents could incorporate different forms of logic in the discretionary viable space, this freedom to take actions granted only a basic level of autonomy to the agents. Such models of agency were also restrictive as there was no learning involved. We next discussed about adaptive agents, based on learning based models. These models have been designed so that agents can learn about taking actions using specific strategies by interacting with the environment and other agents. However, in this case there is no motivation for the agents to choose specific strategies or actions apart from greedily increasing their rewards. We then looked into rational choice and game theoretic models where agents have well defined self-interest functions, and they choose to take actions which maximizes their utility. However, is agency just about self-interest and utility maximization? Is an agent just about its preference function over the action space? Are preference relations arbitrarily defined, or are there underlying foundations that guide an agent's preferences? We addressed this question by involving the concept of self into models of agency. These models posit that actions taken by agents are such that it can maintain a stable state of being (using various self-* properties). The preference functions or its action space are not just about greedy maximization of immediate utility but about prolonging the system to persist in its stable state. 

In our opinion, the concept of self would need to receive increased research attention in order to address deeper elements of intelligence, like general intelligence. There needs to be an intricate model of self for agents which can link their preference and action space to its self. The model of self need not just be the model of self of an individual agent, it can also represent the collective self. 

\bibliographystyle{apacite}
\bibliography{agency}

\end{document}